\documentclass{article}

\usepackage[final, nonatbib]{neurips_2021_ml4ps}

\usepackage[utf8]{inputenc} 
\usepackage[T1]{fontenc}    
\usepackage{hyperref}       
\usepackage{url}            
\usepackage{booktabs}       
\usepackage{amsfonts}       
\usepackage{nicefrac}       
\usepackage{microtype}      
\usepackage{xcolor}         
\usepackage{physics}
\usepackage{graphicx}
\usepackage[sorting=none]{biblatex}
\usepackage{svg}

\newcommand{\E}{\mathbb{E}}
\newcommand{\x}{\mathbf{x}}
\newcommand{\z}{\mathbf{z}}
\newcommand{\X}{\mathbf{X}}
\newcommand{\Z}{\mathbf{Z}}

\definecolor{yescolor}{RGB}{53, 184, 88}
\definecolor{nocolor}{RGB}{204, 167, 57}
\definecolor{nacolor}{RGB}{159, 168, 133}

\title{Turbo-Sim: a generalised generative model with a physical latent space}

\author{%
    Guillaume Quétant$^\text{\normalfont a}{}^\text{\normalfont b}$\thanks{G. Quétant and S. Voloshynovskiy are corresponding authors.} \\
    \texttt{guillaume.quetant@unige.ch} \\
    \And
    Mariia Drozdova$^\text{\normalfont a}{}^\text{\normalfont b}$ \\
    \texttt{mariia.drozdova@unige.ch} \\
    \And
    Vitaliy Kinakh$^\text{\normalfont a}$ \\
    \texttt{vitaliy.kinakh@unige.ch} \\
    \And
    Tobias Golling$^\text{\normalfont b}$ \\
    \texttt{tobias.golling@unige.ch} \\
    \And
    Slava Voloshynovskiy$^\text{\normalfont a}$\footnotemark[1] \\
    \texttt{svolos@unige.ch} \\
    \AND
    \\
    $^\text{a}$Department of Computer Science, University of Geneva, 1227 Carouge \\
    $^\text{b}$Department of Particle Physics, University of Geneva, 1205 Genève
}

\begin{document}

\maketitle

\begin{abstract}
    We present Turbo-Sim, a generalised autoencoder framework derived from principles of information theory that can be used as a generative model. By maximising the mutual information between the input and the output of both the encoder and the decoder, we are able to rediscover the loss terms usually found in adversarial autoencoders and generative adversarial networks, as well as various more sophisticated related models. Our generalised framework makes these models mathematically interpretable and allows for a diversity of new ones by setting the weight of each loss term separately. The framework is also independent of the intrinsic architecture of the encoder and the decoder thus leaving a wide choice for the building blocks of the whole network. We apply Turbo-Sim to a collider physics generation problem: the transformation of the properties of several particles from a theory space, right after the collision, to an observation space, right after the detection in an experiment.
\end{abstract}

\section{Introduction}
\label{sec:intro}

Deep learning models have proven to be promising in tasks consisting in the generation of new data following a desired, sometimes abstract, theoretical model or in the inference of certain parameters of this model given the observed data. Two predominant frameworks in such tasks are adversarial autoencoder (AAE) \cite{Makhzani2015} and generative adversarial network (GAN) \cite{Goodfellow2014}, which have been extensively studied, modified and redesigned in order to fit the problems they are applied to. In this paper we show a new framework, Turbo-Sim, that generalises the aforementioned models and gives a mathematical interpretation to the usual training strategies.

In both AAE and GAN there are two main variable spaces: a data space and a latent space. The key underlying idea of Turbo-Sim is to maximise the mutual information between the two spaces, assuming they are sampled from a joint, usually intractable, probability density. Once the density is carefully parametrized by deep networks, the mutual information can be decomposed and lower bounded in a number of ways, the maximisation of which gives rise to several loss terms, including those of AAEs and GANs, leaving the choice of the exact architecture of the building blocks open.\footnote{A different approach discussed in \cite{Voloshynovskiy2019, Voloshynovskiy2020} gives rise to variational autoencoder (VAE) and links it to GAN.} It also highlights the possibility of training the model in two directions called \textit{direct} and \textit{reverse}, which helps with the pseudo-invertibility of autoencoder-like models. This is expanded in Section~\ref{sec:theory}.

Generative modelling is a crucial step in the process of scientific analysis, and in particular in particle physics. Indeed, the huge amount of data collected from experiments such as the Large Hadron Collider (LHC) needs to be compared with simulated data in order to make statistically meaningful conclusions. The fast growth of data taken by experiments is accompanied by a fast increase of the demand for the generation of simulated data, which is generally the bottleneck of the whole analysis pipeline. In this regard, machine learning aims to be an alternative to classical Monte Carlo generative models. Deep networks might for example be used in replacement of low-level feature simulations such as calorimeter hits \cite{FastCaloGAN2020, AF32021} or energy deposits in detector cells \cite{ATL-SOFT-PUB-2018-001, Buhmann2021}, or of high-level feature simulations such as reconstructed observables from raw detector data \cite{Howard2021}. In Section~\ref{sec:experiment} we focus on the latter and make a direct comparison of our framework with the OTUS method \cite{Howard2021}.\footnote{Code is available at \url{https://github.com/quetant/turbo-sim-supplementary}.}

\section{Turbo-Sim formalism}
\label{sec:theory}

\subsection{Base principle}

\begin{figure}[ht]
    \centering
    \includegraphics[width=\textwidth]{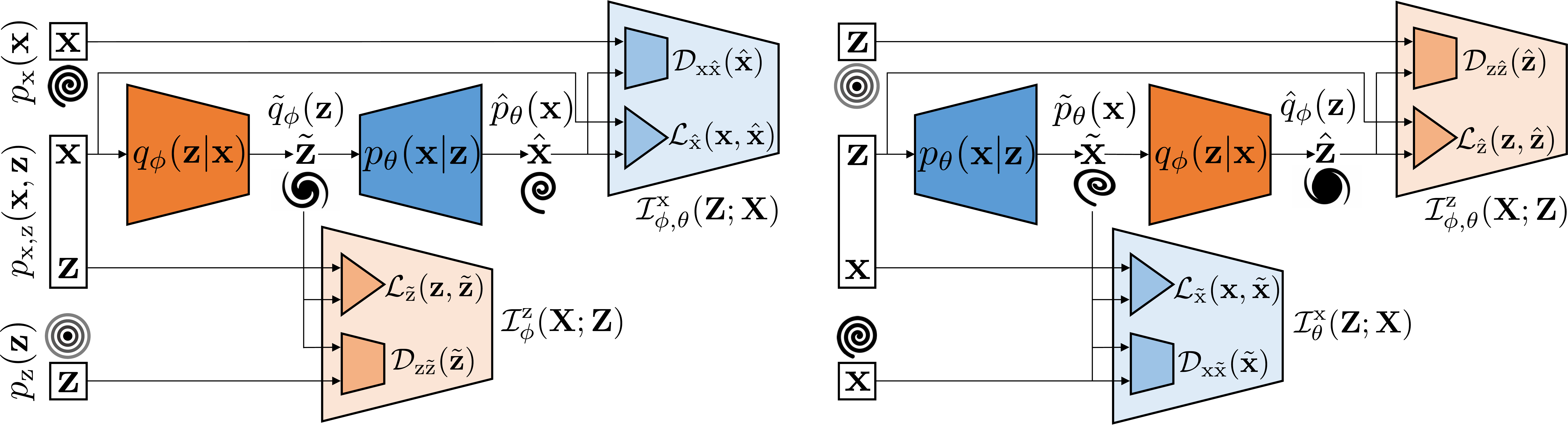}
    \caption{The \textit{direct} (left) and \textit{reverse} (right) directions of the Turbo-Sim framework.}
    \label{fig:turbo}
\end{figure}

Given a dataset composed of pairs $\{\x_i, \z_i\}_{i=1}^N$ sampled from the joint probability density $p_{\mathrm{x}, \mathrm{z}}(\x, \z)$ representing $N$ realisations of the two random variables $\X$ and $\Z$, we show that one can maximise a parametrization of the true mutual information $I(\X; \Z)$ between them in order to get the best approximation. The true mutual information is often intractable since the true joint density $p_{\mathrm{x}, \mathrm{z}}(\x, \z)$ and marginal densities $p_\mathrm{x}(\x)$ and $p_\mathrm{z}(\z)$ are usually unknown. In order to approximate the mutual information, the joint density is first rewritten as:

\begin{equation}
    p_{\mathrm{x}, \mathrm{z}}(\x, \z) = p_{\mathrm{z} | \mathrm{x}}(\z | \x) p_\mathrm{x}(\x) = p_{\mathrm{x} | \mathrm{z}}(\x | \z) p_\mathrm{z}(\z).
\end{equation}

Both conditional densities $p_{\mathrm{z} | \mathrm{x}}(\z | \x)$ and $p_{\mathrm{x} | \mathrm{z}}(\x | \z)$ are then approximated by parametrized ones, $q_\phi(\z | \x)$ and $p_\theta(\x | \z)$ respectively. The marginal densities are denoted as:

\begin{align*}
    \tilde{q}_\phi(\z) &= \E_{p_\x(\x)}\qty[q_\phi(\z|\x)], &
    \hat{p}_\theta(\x) &= \E_{\tilde{q}_\phi(\z)}\qty[p_\theta(\x|\z)], \\
    \tilde{p}_\theta(\x) &= \E_{p_\z(\z)}\qty[p_\theta(\x|\z)], &
    \hat{q}_\phi(\z) &= \E_{\tilde{p}_\theta(\x)}\qty[q_\phi(\z|\x)].
\end{align*}

This leads to multiple ways of decomposing the approximated mutual information.

\subsection{Approximation for the \textit{direct} direction}

After some algebraic manipulations and by approximating $p_{\mathrm{z} | \mathrm{x}}(\z | \x) \approx q_\phi(\z | \x)$, one obtains the two following lower bounds to $I(\X;\Z)$ (details can be found in Appendix~\ref{app:lower_bound_true} and \ref{app:lower_bounds_approx}, as well as in \cite{Voloshynovskiy2019, Voloshynovskiy2020}):

\begin{align}
    &\resizebox{0.94\textwidth}{!}{$\mathcal{I}_\phi^\mathrm{z}(\X; \Z)
    = \E_{p_{\mathrm{x}, \mathrm{z}}(\x, \z)}\qty[\log\frac{q_\phi(\z | \x)}{p_\mathrm{z}(\z)} \frac{\tilde{q}_\phi(\z)}{\tilde{q}_\phi(\z)}]
    \geq \underbrace{\E_{p_\mathrm{x}(\x)}\E_{q_\phi(\z | \x)}\qty[\log q_\phi(\z | \x)]}_{-\mathcal{L}_{\tilde{\mathrm{z}}}(\z, \tilde{\z})} - \underbrace{D_\mathrm{KL}(p_\mathrm{z}(\z) \| \tilde{q}_\phi(\z))}_{\mathcal{D}_{\mathrm{z} \tilde{\mathrm{z}}}(\tilde{\z})}$,} \\
    &\resizebox{0.94\textwidth}{!}{$\mathcal{I}_{\phi, \theta}^\mathrm{x}(\Z; \X)
    = \E_{p_{\mathrm{x}, \mathrm{z}}(\x, \z)}\qty[\log\frac{p_\theta(\x | \z)}{p_\mathrm{x}(\x)} \frac{\hat{p}_\theta(\x)}{\hat{p}_\theta(\x)}]
    \geq \underbrace{\E_{p_\mathrm{x}(\x)}\E_{q_\phi(\z | \x)}\qty[\log p_\theta(\x | \z)]}_{-\mathcal{L}_{\hat{\mathrm{x}}}(\x, \hat{\x})} - \underbrace{D_\mathrm{KL}(p_\mathrm{x}(\x) \| \hat{p}_\theta(\x))}_{\mathcal{D}_{\mathrm{x} \hat{\mathrm{x}}}(\hat{\x})}$.}
\end{align}

One has thus to train the network in such a way as to maximise both $\mathcal{I}_\phi^\mathrm{z}(\X; \Z)$ and $\mathcal{I}_{\phi, \theta}^\mathrm{x}(\Z; \X)$ terms, with a possible weight factor $\alpha$ between them, in order to find the best parameters $\phi$ and $\theta$ of the encoder and the decoder respectively. This is achieved in the \textit{direct} direction by minimising the following loss, leading to the left network shown in Fig.~\ref{fig:turbo}:

\begin{equation}
    \bar{\mathcal{L}}^\mathrm{Direct}(\phi, \theta) = \mathcal{L}_{\tilde{\mathrm{z}}}(\z, \tilde{\z}) + \mathcal{D}_{\mathrm{z} \tilde{\mathrm{z}}}(\tilde{\z}) + \alpha\,\mathcal{L}_{\hat{\mathrm{x}}}(\x, \hat{\x}) + \alpha\,\mathcal{D}_{\mathrm{x} \hat{\mathrm{x}}}(\hat{\x}).
\end{equation}

\subsection{Approximation for the \textit{reverse} direction}

An analogous procedure with the approximation $p_{\mathrm{x} | \mathrm{z}}(\x | \z) \approx p_\theta(\x | \z)$ gives the following two other lower bounds to $I(\X;\Z)$:

\begin{align}
    \mathcal{I}_\theta^\mathrm{x}(\Z; \X)
    &\geq \underbrace{\E_{p_\mathrm{z}(\z)}\E_{p_\theta(\x | \z)}\qty[\log p_\theta(\x | \z)]}_{-\mathcal{L}_{\tilde{\mathrm{x}}}(\x, \tilde{\x})} - \underbrace{D_\mathrm{KL}(p_\mathrm{x}(\x) \| \tilde{p}_\theta(\x))}_{\mathcal{D}_{\mathrm{x} \tilde{\mathrm{x}}}(\tilde{\x})}, \\
    \mathcal{I}_{\phi, \theta}^\mathrm{z}(\X; \Z)
    &\geq \underbrace{\E_{p_\mathrm{z}(\z)}\E_{p_\theta(\x | \z)}\qty[\log q_\phi(\z | \x)]}_{-\mathcal{L}_{\hat{\mathrm{z}}}(\z, \hat{\z})} - \underbrace{D_\mathrm{KL}(p_\mathrm{z}(\z) \| \hat{q}_\phi(\z))}_{\mathcal{D}_{\mathrm{z} \hat{\mathrm{z}}}(\hat{\z})},
\end{align}

and the \textit{reverse} direction loss, weighted by $\beta$, leading to the right network shown in Fig.~\ref{fig:turbo}:

\begin{equation}
    \bar{\mathcal{L}}^\mathrm{Reverse}(\phi, \theta) = \mathcal{L}_{\tilde{\mathrm{x}}}(\x, \tilde{\x}) + \mathcal{D}_{\mathrm{x} \tilde{\mathrm{x}}}(\tilde{\x}) + \beta\,\mathcal{L}_{\hat{\mathrm{z}}}(\z, \hat{\z}) + \beta\,\mathcal{D}_{\mathrm{z} \hat{\mathrm{z}}}(\hat{\z}).
\end{equation}

\subsection{Generic implementation}

The general formalism of the Turbo-Sim framework allows to explain several state-of-the-art autoencoder and generative models. The first crucial point is that the intrinsic encoder and decoder architectures are not specified. Indeed they could be any deterministic or stochastic mappers, $q_\phi(\z | \x)$ and $p_\theta(\x | \z)$ respectively. Dense network might be used for vector data, convolutional network for images, graph network for point cloud data, or any more sophisticated architecture for a given specific data structure. The second point is that by turning on or off certain weights ($\alpha$, $\beta$, $\gamma$, ...) of the general loss:

\begin{equation}
    \bar{\mathcal{L}}^\mathrm{Turbo}(\phi, \theta) = \bar{\mathcal{L}}^\mathrm{Direct}(\phi, \theta) + \gamma\,\bar{\mathcal{L}}^\mathrm{Reverse}(\phi, \theta),
\end{equation}

one can recover a wide range of common training strategies. Here we allow a factor $\gamma$ between the two losses to weight them independently. As a simple example, standard AAE \cite{Makhzani2015} is described by $\mathcal{D}_{\mathrm{z}, \tilde{\mathrm{z}}}(\tilde{\z})$ approximated by a discriminator network on the latent space and $\mathcal{L}_{\hat{\mathrm{x}}}(\x, \hat{\x})$ given by the supervised $\ell_p$-norm on the data space assuming an exponential distribution for $p_\theta(\x | \z) \propto \exp(-\lambda \|\x - \hat{\x}\|_p^p)$, where $\hat{\x}$ is a function of $\z$. Notice that thanks to the $\alpha$, $\beta$ and $\gamma$ weights and because of the $\lambda$ factors found in exponential distributions, all eight terms of $\bar{\mathcal{L}}^\mathrm{Turbo}(\phi, \theta)$ receive an independent factor and thus can be weighted independently.



\section{A generative model for particle physics}
\label{sec:experiment}

\begin{figure}[ht]
    \centering
    \includegraphics[height=0.2\textheight]{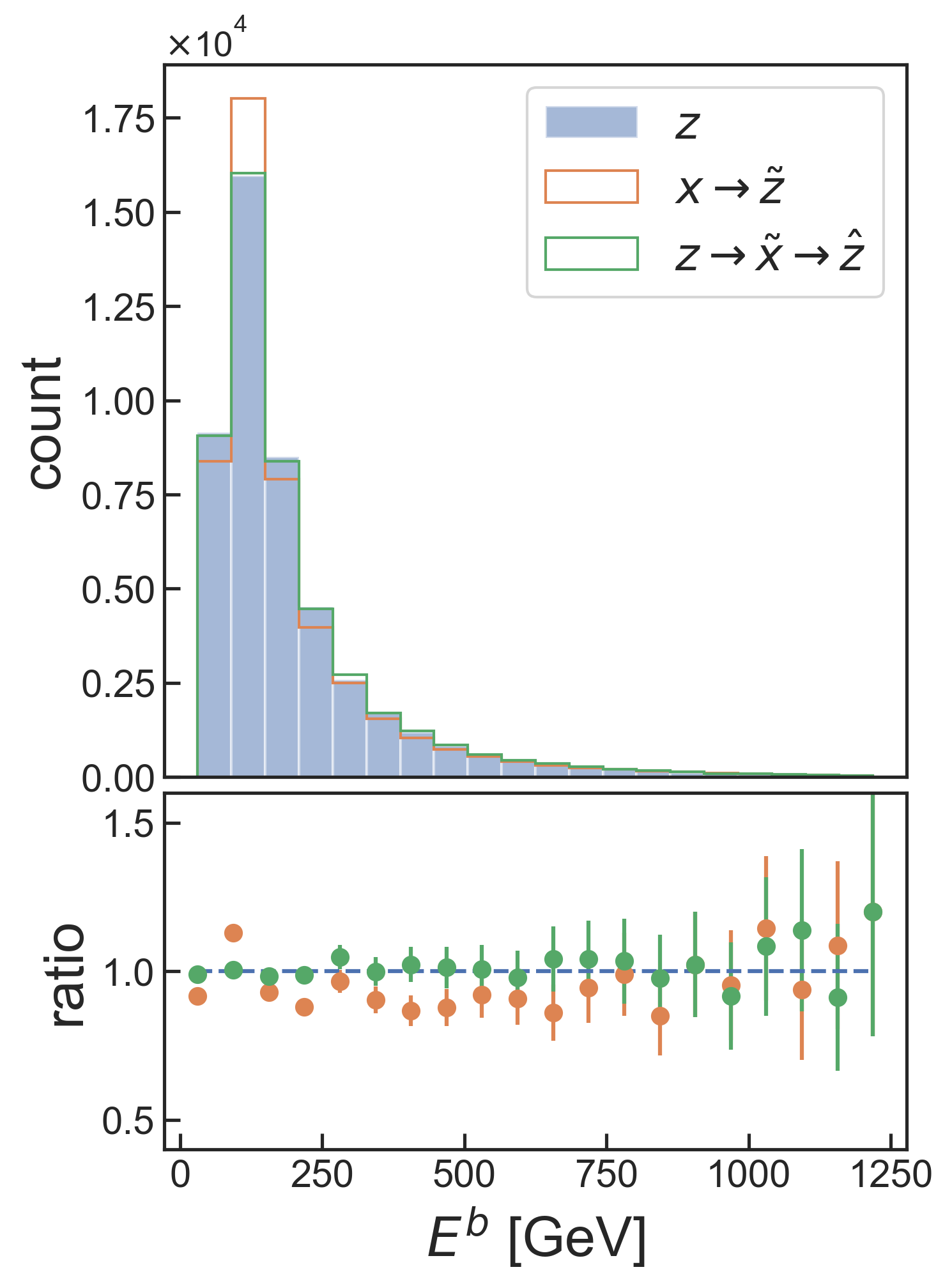}
    \includegraphics[height=0.2\textheight]{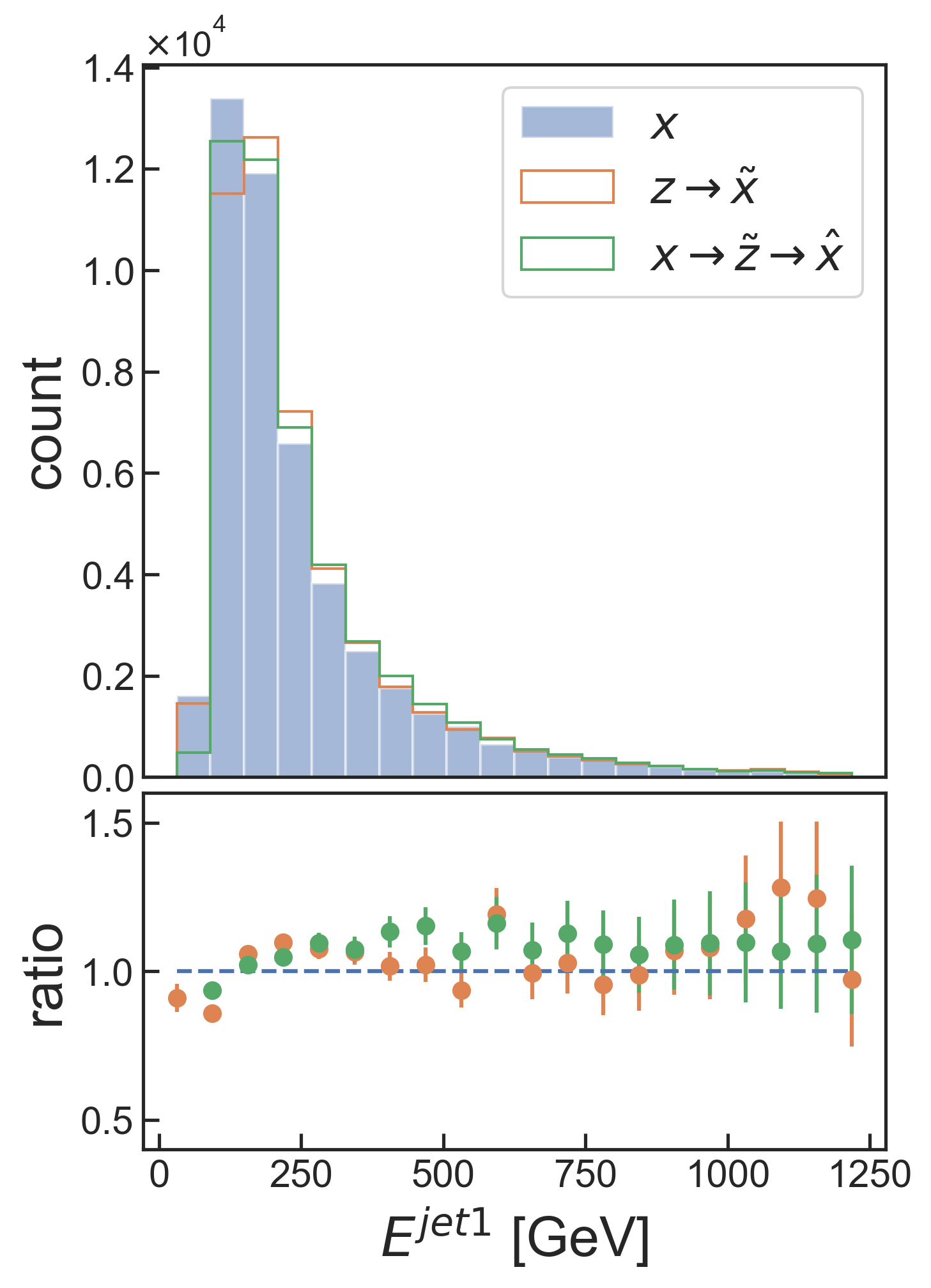}
    \includegraphics[height=0.2\textheight]{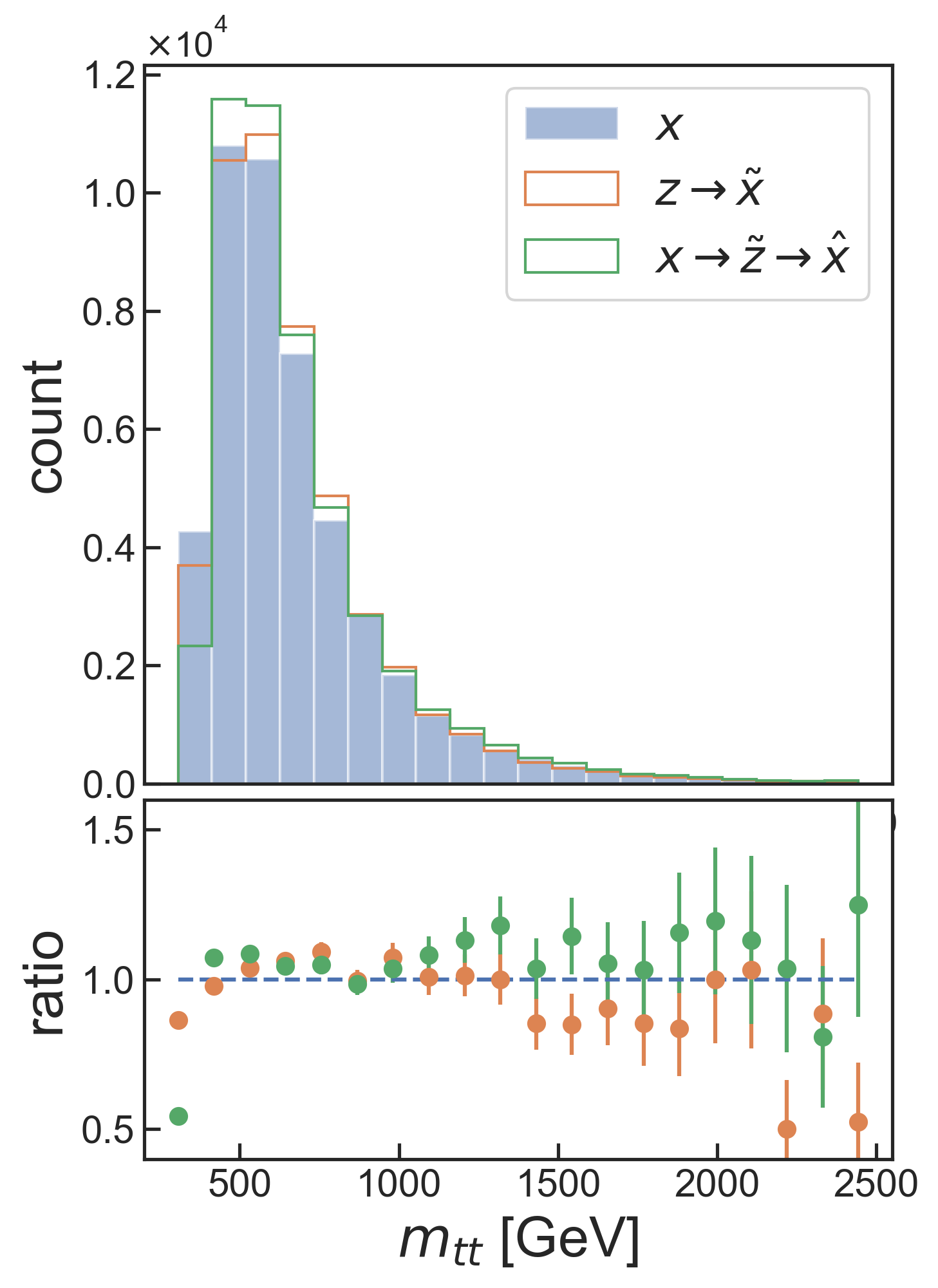}
    \includegraphics[height=0.2\textheight]{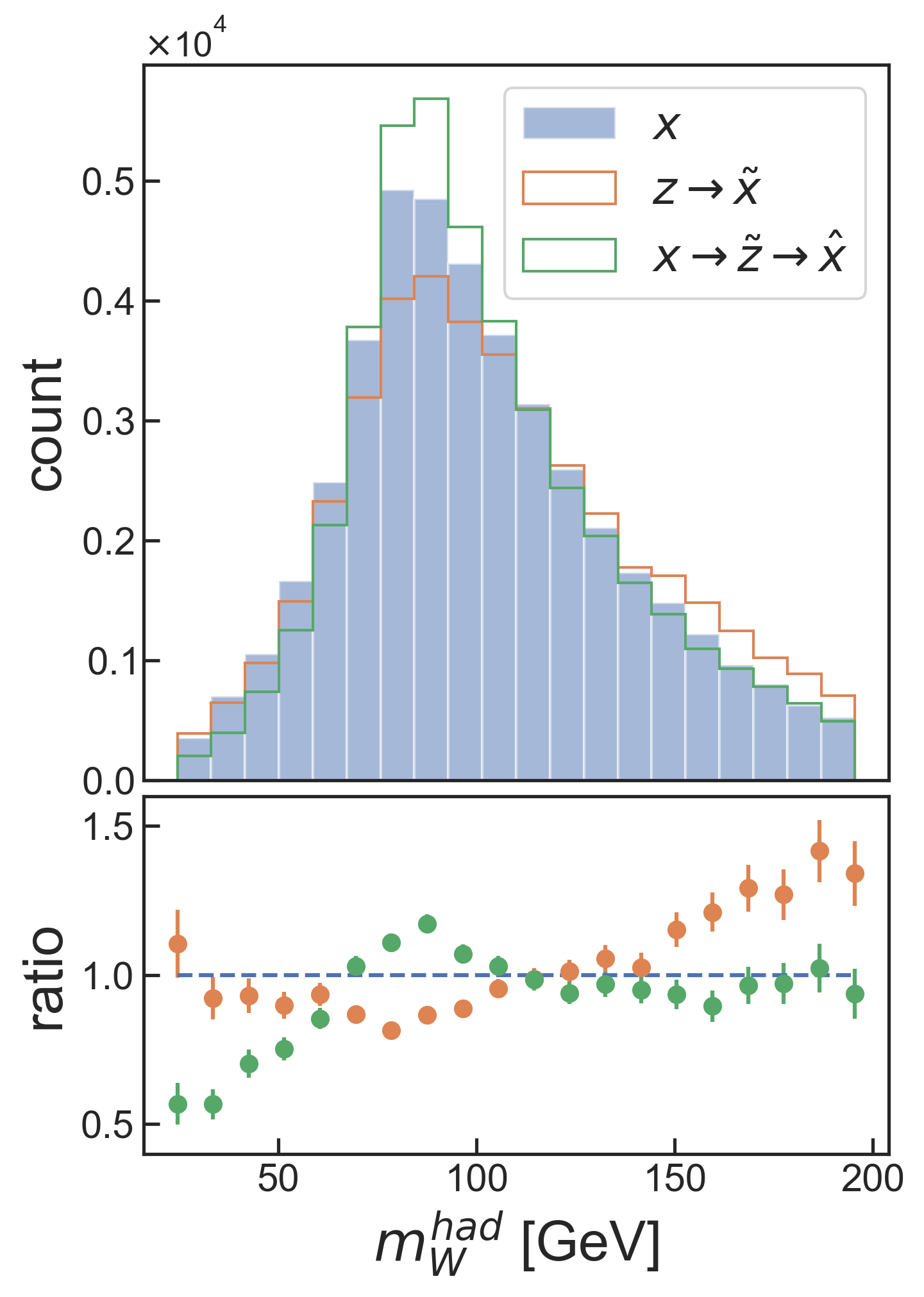}
    \caption{Distributions of a selection of observables for truth (blue), sampled (orange) and reconstructed (green) data.}
    \label{fig:physics}
\end{figure}

We apply our generalised Turbo-Sim framework to a collider physics transformation problem. In particular, we will focus on double top quarks production in proton--proton collision subsequently decaying into b-quarks and W-bosons which in turn decay into a pair of light leptons and a pair of light quarks, $pp \rightarrow t\bar{t} \rightarrow e^-\bar{\nu}_e \, b\bar{b} \, u\bar{d}$ as described in \cite{Howard2021}.\footnote{The dataset was shared with us and is now publicly available at \url{https://github.com/yiboyang/otus}.} The final state of this process is thus made of six particles whose four-momenta are taken to form the $\Z$ space. After parton shower, hadronisation and potential formation of jets in the detectors, the four-momenta assigned to the reconstructed objects are taken to form the $\X$ space. In this space, one loses information about the original particles, due to mainly two reasons. Firstly, the neutrino cannot be detected, hence it is indirectly defined by the missing transverse momentum, which must sum up to zero. Secondly, the four jets cannot be assigned to the four initial quarks since they all decay in statistically identical showers in the detector. They are thus treated as independent objects.

Transforming the random distributions of observables well defined by the rules of quantum field theory into somewhat modified distributions of the same observables due to the stochastic nature of the experiments is a common problem in particle physics generative modelling. The key point in this scenario is the word \textit{transformation}. Indeed, standard simulation tools take the $\Z$ space as input and transform it by applying physically motivated stochastic processes in order to output the $\X$ space \cite{Alwall2014, Gean42003, Favereau2014}. On the other hand, usual deep learning models are trained to mimic the same outputs by taking Gaussian, or any tractable distributions, as input. The paradigm shift introduced in \cite{Howard2021} is to rather take this $\Z$ space as latent space for the deep network, and thus as input for the generative model. We follow the same line to train the full Turbo-Sim network shown in Fig.~\ref{fig:turbo} and select the best model after a random hyperparameters search.\footnote{The random hyperparameters search was performed on an internal cluster with either nVidia Titan~X, P100, RTX~2080~Ti, RTX~3090 or A100 GPUs, for a maximum of 12 hours.} Details can be found in the code.

The distributions of a selection of observables resulting from our trained Turbo-Sim model are shown in Fig.~\ref{fig:physics}, where sampled ($\tilde{\z}$ and $\tilde{\x}$) and reconstructed ($\hat{\z}$ and $\hat{\x}$) data are compared to the truth ($\z$ and $\x$). A quantitative comparison of samples with the OTUS method is shown in Tab.~\ref{tab:ks} by means of the Kolmogorov-Smirnov test, which defines a distance between two cumulative distributions. $p_y^b$, $p_z^b$, $E^b$ are the momenta and energy of the b-quark, $p_y^{jet1}$, $p_z^{jet1}$, $E^{jet1}$ are the momenta and energy of the leading jet. These variables, among others, are used during the training. From them we compute the masses of several underlying physical objects in order to verify how well the underlying physical correlations are learnt. $m_{tt}$ is the top quark pair mass, $m_W^{had}$ is the hadronic W-boson mass, $m_t^{lep}$, $m_t^{had}$ are the leptonic and hadronic top quark masses. The reconstructed W-boson and top quark masses are computed from the four-momentum of the four jets, the electron and the reconstructed neutrino in the $\X$ space. All proper combinations of objects are compared to the true masses thanks to a chi-square test, the best one being kept. The neutrino longitudinal momentum is computed by solving $m_W^2 = (p^{e^-} + p^\nu)^2$.

We observe that, although the OTUS method looks great at sampling the $\Z$ space, the Turbo-Sim method is better at sampling the $\X$ space, which is the main goal of a generative model. In addition, Turbo-Sim outperforms OTUS when it comes to the reconstruction of decayed objects, implying that the underlying physics has been better learnt.


\begin{table}[ht]
  \caption{Kolmogorov-Smirnov distance to truth [$\times 10^{-2}$] for several observables including those of Fig.~\ref{fig:physics}. A lower value means a higher accuracy.}
  \label{tab:ks}
  \centering
  \begin{tabular}{lllllllllll}
    \toprule
                & \multicolumn{3}{c}{$\mathbf{Z}$ space} & \multicolumn{3}{c}{$\mathbf{X}$ space}   & \multicolumn{4}{c}{Reconstructed physics}               \\
                \cmidrule(l){2-4}                   \cmidrule(l){5-7}                               \cmidrule(l){8-11}
    Model       & $p_y^b$   & $p_z^b$   & $E^b$     & $p_y^{jet1}$  & $p_z^{jet1}$  & $E^{jet1}$    & $m_{tt}$  & $m_W^{had}$   & $m_t^{lep}$   & $m_t^{had}$ \\
    \midrule
    Turbo-Sim       & 5.28      & 7.28      & 3.96      & \bf{2.89}     & 10.3          & \bf{4.43}     & \bf{2.97} & \bf{7.72}     & \bf{5.20}     & \bf{8.52}   \\
    \midrule
    OTUS        & \bf{1.59} & \bf{1.23} & \bf{2.76} & 3.78          & \bf{2.39}     & 5.75          & 15.8      & 11.7          & 14.1          & 24.9        \\
    \bottomrule
  \end{tabular}
\end{table}

\section{Discussion}
\label{sec:conclusion}

In this work we presented a new autoencoder interpretation based on the maximisation of the mutual information between the latent space and the data space. We used this formalism to build a generative model for particle physics called Turbo-Sim and showed that it is able to compete with state-of-the-art methods such as OTUS, even outperforming it in critical tasks. It should be emphasised that these results are reached with very basic network building blocks: both the encoder and the decoder are only made of fully connected layers, as well as the Wasserstein gradient penalty discriminators. Future implementations with more expressive networks are expected to enhance the performance. One key improvement would be in the way stochasticity is handled. Instead of simply adding some noise to the input of the networks, one might use a layer-wise noise implementation following the EigenGAN architecture \cite{He2021}. This way, one allows the model to learn how to apply different noise inputs to different hidden features of the data, thus controlling them separately. Moreover, in contrast with the OTUS method, we did not include any physically motivated constraints such as momentum alignment or mass conservation alongside our Turbo-Sim loss terms, while yielding comparable performance.

Another interesting application of such models is to use them for unfolding tasks. Indeed, an important aspect of particle physics analysis is also to get back from the observed data to the actual physics. In our language, this means going from the $\X$ space to the $\Z$ space. Thanks to the paradigm of using a physically meaningful latent space, i.e. the theoretical distributions of energy and momenta, it turns out that our Turbo-Sim model is also trained to achieve this task, but its performance is still below that of OTUS. However, because of the symmetrical nature of the network that manifests itself with the \textit{direct} and \textit{reverse} direction of training, it should be able to also learn this transformation properly. Nevertheless, a limitation could come from the data itself. Due to the highly stochastic nature of detector observations, the $\X$ space shows more smoothed out features than the $\Z$ space, which makes the inference direction harder. Future EigenGAN-like implementations might help handling this stochasticity, hence we are confident that a more expressive model will be able to perform equally well in both generation and unfolding tasks.

\begin{ack}

We would like to thank Sebastian Pina-Otey and Johnny Raine for fruitful discussions, as well as for their enlightening comments and feedback. This research was partially funded by the SNF Sinergia project (CRSII5\_193716): Robust Deep Density Models for High-Energy Particle Physics and Solar Flare Analysis (RODEM).

\end{ack}

\printbibliography

\newpage
\appendix

\section{Appendix}

\subsection{Lower bounds to the true mutual information}
\label{app:lower_bound_true}

In this section, we prove that the parametrization of the true mutual information between the $\X$ and the $\Z$ spaces $I(\X;\Z) \simeq \mathcal{I}_\phi^\z(\X;\Z)$ is not only an approximation, but also a lower bound i.e. $I(\X;\Z) \geq \mathcal{I}_\phi^\z(\X;\Z)$:

\begin{align}
    I(\X;\Z)
    &= \E_{p_{\x, \z}(\x, \z)} \qty[\log\frac{p_{\x, \z}(\x, \z)}{p_{\x}(\x) p_{\z}(\z)}] \nonumber \\
    &= \E_{p_{\x, \z}(\x, \z)} \qty[ \log\frac{p_{\z | \x}(\z | \x)}{p_{\z}(\z)}] \nonumber \\
    &= \E_{p_{\x, \z}(\x, \z)} \qty[ \log\frac{p_{\z | \x}(\z | \x)}{p_{\z}(\z)} \frac{q_\phi(\z | \x)}{q_\phi(\z | \x)}] \nonumber \\
    &= \E_{p_{\x, \z}(\x, \z)} \qty[ \log\frac{q_\phi(\z | \x)}{p_{\z}(\z)}] + \E_{p_{\x, \z}(\x, \z)} \qty[ \log\frac{p_{\z | \x}(\z | \x)}{q_\phi(\z | \x)}] \\
    &= \E_{p_{\x, \z}(\x, \z)} \qty[ \log\frac{q_\phi(\z | \x)}{p_{\z}(\z)}] + \E_{p_\x(\x)} \qty[ D_\mathrm{KL}(p_{\z | \x}(\z | \x) \| q_\phi(\z | \x))] \nonumber \\
    &\geq \E_{p_{\x, \z}(\x, \z)} \qty[ \log\frac{q_\phi(\z | \x)}{p_{\z}(\z)}] \nonumber \\
    &= \E_{p_{\x, \z}(\x, \z)} \qty[ \log\frac{q_\phi(\z | \x)}{p_{\z}(\z)} \frac{\tilde{q}_\phi(\z)}{\tilde{q}_\phi(\z)}] = \mathcal{I}_\phi^\z(\X;\Z), \nonumber
\end{align}

where the inequality holds because $D_\mathrm{KL}(p_{\z | \x}(\z | \x) \| q_\phi(\z | \x)) \geq 0$ and where the equality is reached if and only if $p_{\z | \x}(\z | \x) = q_\phi(\z | \x)$. Maximising the approximated  mutual information ensures that the parametrized density $q_\phi(\z | \x)$ is as close as possible to the true density $p_{\z | \x}(\z | \x)$. The proves for the three other lower bounds $\mathcal{I}_{\phi, \theta}^\mathrm{x}(\Z; \X)$, $\mathcal{I}_\theta^\mathrm{x}(\Z; \X)$ and $\mathcal{I}_{\phi, \theta}^\mathrm{z}(\X; \Z)$ follow the exact same logic with:

\begin{align*}
    \mathcal{I}_{\phi, \theta}^\mathrm{x}(\Z; \X) &= \E_{p_{\mathrm{x}, \mathrm{z}}(\x, \z)}\qty[\log\frac{p_\theta(\x | \z)}{p_\mathrm{x}(\x)} \frac{\hat{p}_\theta(\x)}{\hat{p}_\theta(\x)}] \\ \mathcal{I}_\theta^\mathrm{x}(\Z; \X) &= \E_{p_{\mathrm{x}, \mathrm{z}}(\x, \z)}\qty[\log\frac{p_\theta(\x | \z)}{p_\mathrm{x}(\x)} \frac{\tilde{p}_\theta(\x)}{\tilde{p}_\theta(\x)}] \\
    \mathcal{I}_{\phi, \theta}^\mathrm{z}(\X; \Z) &= \E_{p_{\mathrm{x}, \mathrm{z}}(\x, \z)}\qty[\log\frac{q_\phi(\z | \x)}{p_\mathrm{z}(\z)} \frac{\hat{q}_\phi(\z)}{\hat{q}_\phi(\z)}]
\end{align*}

\subsection{Lower bounds to the approximate mutual information}
\label{app:lower_bounds_approx}

In this section, we derive a lower bound to the parametrized approximation of the mutual information $\mathcal{I}_\phi^\mathrm{z}(\X; \Z)$ by decomposing it for the \textit{direct} direction as:

\begin{equation}
\begin{aligned}
    \mathcal{I}_\phi^\mathrm{z}(\X; \Z)
    &= \E_{p_{\mathrm{x}, \mathrm{z}}(\x, \z)}\qty[\log\frac{q_\phi(\z | \x)}{p_\mathrm{z}(\z)} \frac{\tilde{q}_\phi(\z)}{\tilde{q}_\phi(\z)}] \\
    &= \E_{p_{\mathrm{x}, \mathrm{z}}(\x, \z)}[\log q_\phi(\z | \x)] - \E_{p_{\mathrm{x}, \mathrm{z}}(\x, \z)}\qty[\log\frac{p_\mathrm{z}(\z)}{\tilde{q}_\phi(\z)}] - \E_{p_{\mathrm{x}, \mathrm{z}}(\x, \z)}[\log \tilde{q}_\phi(\z)] \\
    &= \E_{p_\mathrm{x}(\x)}\E_{p_{\mathrm{z} | \mathrm{x}}(\z | \x)}[\log q_\phi(\z | \x)] - \E_{p_\mathrm{z}(\z)}\qty[\log\frac{p_\mathrm{z}(\z)}{\tilde{q}_\phi(\z)}] - \E_{p_\mathrm{z}(\z)}[\log \tilde{q}_\phi(\z)] \\
    &= \E_{p_\mathrm{x}(\x)}\E_{p_{\mathrm{z} | \mathrm{x}}(\z | \x)}\qty[\log q_\phi(\z | \x)] - D_\mathrm{KL}(p_\mathrm{z}(\z) \| \tilde{q}_\phi(\z)) + H(p_\mathrm{z}(\z); \tilde{q}_\phi(\z)) \\
    &\geq \E_{p_\mathrm{x}(\x)}\E_{p_{\mathrm{z} | \mathrm{x}}(\z | \x)}\qty[\log q_\phi(\z | \x)] - D_\mathrm{KL}(p_\mathrm{z}(\z) \| \tilde{q}_\phi(\z)) \\
    &\simeq \E_{p_\mathrm{x}(\x)}\E_{q_\phi(\z | \x)}\qty[\log q_\phi(\z | \x)] - D_\mathrm{KL}(p_\mathrm{z}(\z) \| \tilde{q}_\phi(\z)),
\end{aligned}
\end{equation}

where the inequality holds because $H(p_\mathrm{z}(\z); \tilde{q}_\phi(\z)) = H(p_\z(\z)) + D_\mathrm{KL}(p_\z(\z) \| \tilde{q}_\phi(\z)) \geq 0$. The equality is not reached for $p_{\z | \x}(\z | \x) = q_\phi(\z | \x)$ because the cross-entropy does not reduce to zero but to the entropy of the prior on $\z$. However, the maximum value of the right-hand side of the inequality is still reached for this precise case, as otherwise the value of the left-hand side could exceed its upper bound described in Appendix~\ref{app:lower_bound_true}. Analogous derivations can be performed for the three remaining approximations $\mathcal{I}_{\phi, \theta}^\mathrm{x}(\Z; \X)$, $\mathcal{I}_\theta^\mathrm{x}(\Z; \X)$ and $\mathcal{I}_{\phi, \theta}^\mathrm{z}(\X; \Z)$ whose expressions can be found in Appendix~\ref{app:lower_bound_true}.

\end{document}